# FedGrad: Optimisation in Decentralised Machine Learning

Mann Patel
CSE DEPSTAR, CHARUSAT
Vadodara, India
18dcs074@charusat.edu.in

*Abstract*—**Federated Learning is a machine learning paradigm where we aim to train machine learning models in a distributed fashion. Large number of clients/edge devices collaborate with each other to train a single model on the central. Clients do not share their own dataset with each other, decoupling computation and data on the same device.**

**In this paper, we propose yet another adaptive federated optimization method, and some other ideas in the field of federated learning. We also perform experiments using these methods and showcase the improvement in overall performance of federated learning.**

## I. Introduction

In reality, data is not identically distributed across multiple clients that want to preserve their data's privacy. To tackle this challenge Federated Learning[4] paradigm was introduced to work on client's private data. Edge device's collaboratively train a model on the central server by sharing 'local updates. On a bigger picture this can be achieved by performing 3 steps, reiteratively: (i) edge device's share local updates of parameters to each other, (ii) central server aggregates the local parameter updates from multiple edge devices, (ii) the aggregated updates are then performed on the shared model and the model is sent back to each edge device for next round.

Usage of orthodox optimization methods like SGD in a distributed fashion is incompatible with our FL setup due to high communication cost. To overcome that, edge devices would locally update themselves multiple times before sharing the update to the central server for aggregation. This greatly reduced communication cost. Very popular optimization method adopting this technique was FEDAVG[4]. Furthermore, after local updates were received by the central server. The updates were aggregated by simply averaging them, to get new shared model. Though the method gained fame, it faced some serious issues : (i) convergence[10] (ii) and lack of adaptability.

In this paper, we present yet another adaptive optimization method called FEDGRAD for federated setting which tackles the second issue of adaptability.

## II. Related work

FEDAVG was first introduced by McMahan et al. (2017), who showed it can dramatically reduce communication costs. Many variants have since been proposed to tackle issues such as convergence and client drift. Examples include adding a regularization term in the client objectives towards the broadcast model (Li et al., 2018), and server momentum (Hsu et al., 2019). Adaptive federated optimizations were first introduced by Reddi et al.(2021) which provided a general framework to design cross-device optimization methods in a nonconvex setting. With this general framework they also published 3 adaptive optimization methods namely, (i) FEDADAGRAD (ii) FEDYOGI (iii) FEDADAM. The proposed methods have the same communication cost as of FEDAVG[4] but converges much faster than FEDAVG, O( $1/\sqrt{mKT}$ ).

## III. Privacy

Albeit the first information isn't traded in FL, the model boundaries can likewise release touchy data with regards to the preparation information. Hence, give protection assurances to the traded neighborhood refreshes.

Differential protection is a well-known strategy to give security ensures. Geyer et al. apply differential protection in combined averaging from a customer level point of view. They utilize the Gaussian component to contort the number of updates of angles to ensure an entire customer's dataset rather than a solitary information point.

McMahan et al. send combined averaging in the preparation of LSTM. They additionally use customer level differential security to ensure the boundaries. Bhowmick et al. apply neighborhood differential protection to ensure the boundaries in FL. To build the model quality, they consider a viable danger model that wishes to decipher people's information yet has minimal earlier data on them. Inside this supposition, they can all the more likely use the protection financial plan.

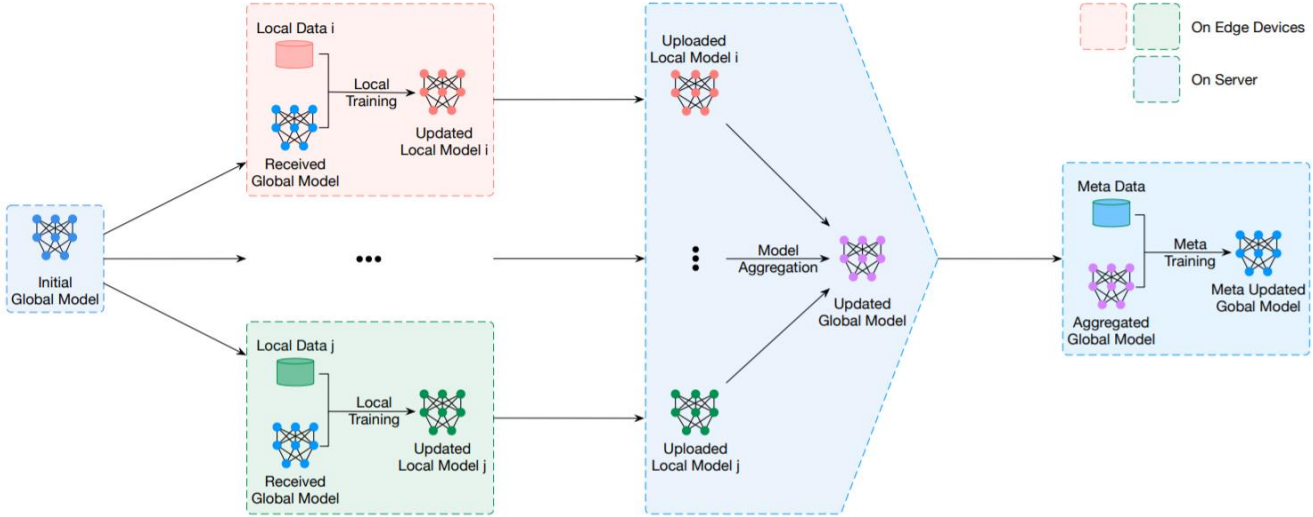

*Figure 1* Federated learning workflow using federated optimizations, then aggregation and model global model update.

Bonawitz et al. apply secure multi-party calculation to ensure the neighborhood boundaries on the premise of united averaging. In particular, they present a solid total convention to safely register the amount of vectors dependent on secret sharing . They additionally talk about how to join differential security with secure conglomeration.

Truex et al. consolidate both secure multiparty calculation and differential privacy for privacy preserving FL. They utilize differential security to infuse clamors to the nearby updates. Then, at that point, the boisterous updates will be encoded utilizing the Paillier cryptosystem before shipped off the focal server. For the assaults on FL, one sort of famous assault is indirect access assault, which intends to accomplish an awful global model by trading malignant nearby updates.

| Networking Protocol | Focus | Benefit(s) |
|---|---|---|
| Hybrid FL | Communication Resource Scheduling Accuracy | Accuracy |
| FedCS | Client's training process | Higher accuracy Robust Models |
| PrivFL | Mobile Networks Data Privacy | Accounts for threats |
| VerifyNet | User Privacy Integrity | Receptiveness High Security |
| FedGRU/JAP | Traffic flow prediction | Reduce overhead |

**Table 1** Protocol based Optimization techniqes for fedarated settings.

## IV. FEDARATED LEARNING SETTING

Fedarated setting differs from other usual distributed settings by 3 major properties:

- **Unbalanced:** The data in unfederated has high likelihood of being unbalanced in the sense that some users will have more data while others may have presumably less data for client-side training
- **NON IID:** The local training data of the client isn't identical and independent as compared to others.

Thus, a client's training data isn't demonstrative of the whole population.

- **Unreliable communication:** The client's device, generally mobile devices are bound to have irregular in having continuous connection with the server. Furthermore, an idle state of device is required due to high CPU/GPU usage of mobile device. Parameter passing is also costly or inefficient.

We deal with a non-convex optimization problem that is just generalization for a single client training, but for multiple clients.

$$\min_{x \in \mathbb{R}^d} f(x) = \frac{1}{m} \sum_{i=1}^{m} F_i(x)$$

Where $F_i(x)$ is loss function of the ith client with data distribution $Ez \sim Di$ and $i \neq j$, meaning Client $i$ will have unidentical data distribution than Client $j$.

$$x_{t+1} = \frac{1}{|\mathcal{S}|} \sum_{i \in \mathcal{S}} x_i^t = x_t - \frac{1}{|\mathcal{S}|} \sum_{i \in \mathcal{S}} (x_t - x_i^t)$$

A very sound, and common approach is to take averages of all the models and update the global model. To make the communication efficient, we sample updates from only a few clients. This approach is called FEDAVG (McMahan et al.), which performed better than FEDSGD, which was costly in terms of computation and while communication as well.

We follow a template introduce by Sashank et al. for any adaptive federated optimizer as mentioned below.

| *Algorithm 1: Adaptive Federated Optimizer template* |
|---|
| 1: Input $M_0$, $OPTIMIZER_{client}$, $OPTIMIZER_{server}$ |
| 2: for r = 0….R − 1 do |
| 3:    Sample $K$ clients |
| 4:    $M_{i,0} = M_r$ |
| 5:    for each client in K, in parallel do |

```
6:      compute predictions Yᵢ from Mᵢ
7:      Δ_c = Mᵢ - OPTIMIZER_client(Mᵢ, Yᵢ, lr, r)
8:      Δ_s = (1/|K|) Σ_{i∈K} Δᵢʳ
9:      M_{r+1} = OPTIMIZER_server(Mᵢ, Δ_s, lr, r)
```

## V. EXPERIMENTS AND ANALYSIS

We compare different hyperparameters in federated setting. We use FLWR framework (Beutel et al) to perform our experiments on the same machine, but on different ports. We work on CIFAR dataset and distribute a simple Convolution neural network to mutually train the server/global model while training locally. We apply some augmentations to make CIFAR dataset, NON-IID.

First let us have 4 clients, with sampling ratio as 0.5 and train the global model using FEDAVG as our server optimizer. We use SGD as our client optimizer, using 0.001 as our client learning rate and momentum $m = 0.9$.

| Task | ADAGRAD | ADAM | YOGI | AVGM | AVG |
|------|---------|------|------|------|-----|
| CIFAR-10 | 72.1 | 77.4 | **78.4** | 77.4 | 72.8 |
| CIFAR-100 | 47.9 | **52.5** | 52.4 | 52.4 | 44.7 |
| EMNIST CR | 85.1 | **85.6** | 85.5 | 85.2 | 84.9 |
| SHAKESPEARE | **57.5** | 57.0 | 57.2 | 57.3 | 56.9 |
| SO LR | **67.1** | 65.8 | 65.9 | 36.9 | 30.0 |
| EMNIST AE | 4.20 | 1.01 | **0.98** | 1.65 | 6.47 |

From the above result we find that, adaptive optimizers perform better in most of the tasks, setting 100 rounds to be the standard.

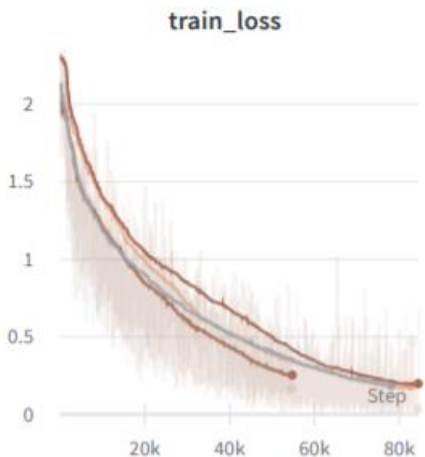

**Figure 2:** Train loss of the CIFAR10 task: 4 devices, 100 rounds, FedAvg as optimizer.

Using this standardized testing we compare different optimizers; we find out adaptive optimizers converge faster than other trivial optimizers.

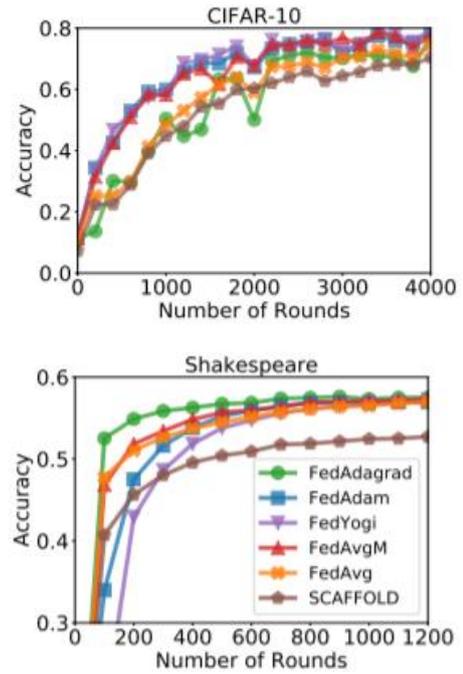

**Figure 3** Results of standard tasks with different optimizers.

To test scalability of federated learning setup, we test the same task with different number of clients participating.

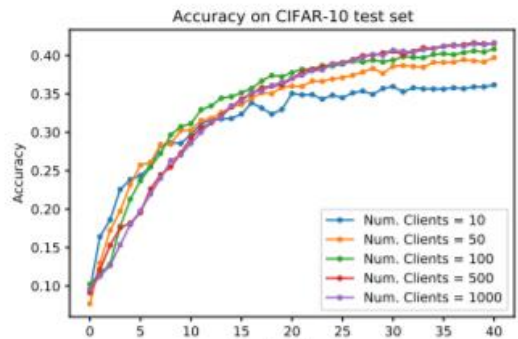

**Figure 4** Accuracy on CIFAR10 dataset v/s number of rounds. As we increase number of clients accuracy increases.

## VI. IMPROVING COMMUNICATION EFFICIENCY

**Structured Updates** This type of communication efficient update by using low rank matrix transformation of updates and random masks on the matrix to make it sparse, having unique masks every round.

**Sketched Updates** In this type of communication efficient updates we compute the update and then compress the updates which can be decompressed by the server upon receiving the updates. Some techniques of compression/encoding are: 1) Subsampling 2) Probablistic Quantization 3) Quantization with structured random rotations.

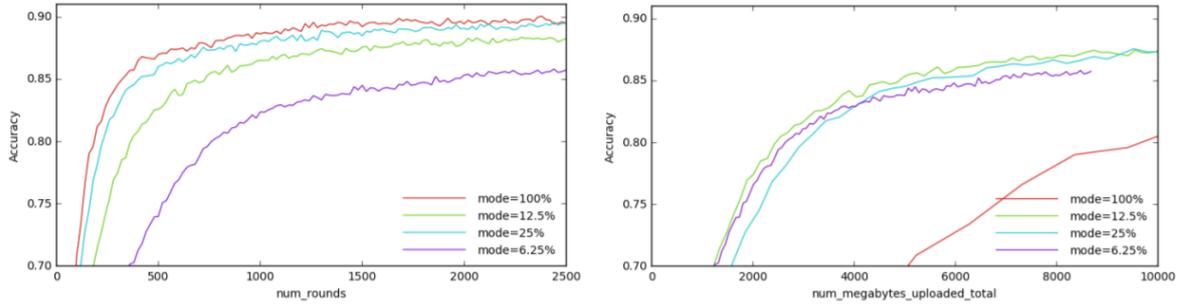

**Figure 6** CIFAR10 task with structured updates for low communication cost using *Low Rank technique*

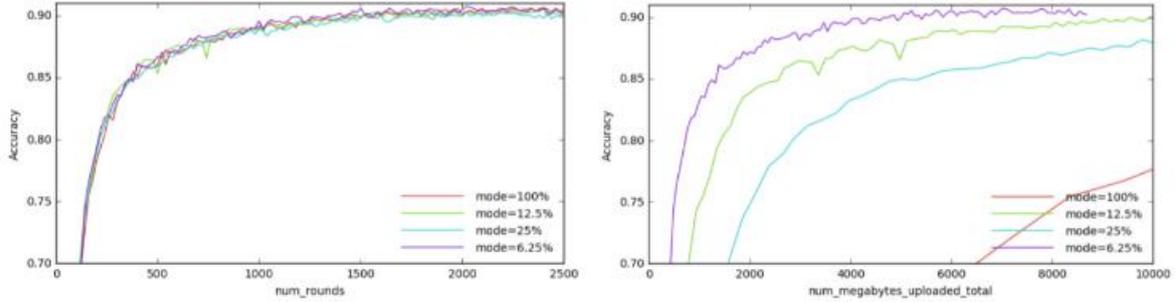

**Figure 5** CIFAR10 task with structured updates for low communication cost using *Random mask technique*

## VII. SPEED HETEROGENEITY

***Heterogenous network speeds across clients:*** We setup an experiment where different client has different network speeds sampled for 3G(normalized download speed ~ 7Mbps) and 4G(normalized download speed~40Mbps) network.

There are two vital focal points from this investigation: a) We can profile the preparation season of any FL calculation under situations of organization heterogeneity, b) we can use these bits of knowledge to configuration refined customer inspecting strategies. For instance, during ensuing rounds of combined learning, we could screen the quantity of tests every customer had the option to process during a given time window and increment the determination likelihood of slow customers to adjust the commitments of quick and slow customers to the global model.